# GANDiff-FR: Hybrid GAN–Diffusion Synthesis for Causal Bias Attribution in Face Recognition *

Md Asgor Hossain Reaj
*Department of Computer Science*
*AIUB*, Dhaka, Bangladesh
20-43999-2@student.aiub.edu

Rajan Das Gupta
*Department of Computer Science*
*AIUB*, Dhaka, Bangladesh
18-36304-1@student.aiub.edu

Md Yeasin Rahat
*Department of Computer Science*
*AIUB*, Dhaka, Bangladesh
20-43097-1@student.aiub.edu

Nafiz Fahad
*Faculty of Information Science and Technology*
*Multimedia University*, Melaka, Malaysia
fahadnafiz1@gmail.com

Md Jawadul Hasan
*Elite Research Lab*
Queens, NY, United States
mdjawadulhasan@gmail.com

Tze Hui Liew
*Centre for Intelligent Cloud Computing*
*Faculty of Information Science and Technology*,
*Multimedia University*, Melaka, Malaysia
thliew@mmu.edu.my

***Abstract*—Bias in face recognition systems can lead to real-world harms—from wrongful arrests to exclusion from essential services—yet current benchmarks cannot isolate the root causes. We introduce GANDiff-FR, the first synthetic framework that precisely controls demographic and environmental factors to measure, explain, and reduce bias with reproducible rigor. GANDiff-FR unifies StyleGAN3-based identity-preserving generation with diffusion-based attribute control, enabling fine-grained manipulation of pose (±30°), illumination (four directions), and expression (five levels) under ceteris paribus conditions. We synthesize 10,000 demographically balanced faces across five cohorts—validated for realism via automated detection (98.2%) and human review (89%)—to isolate and quantify bias drivers. Benchmarking ArcFace, CosFace, and AdaFace under matched operating points shows AdaFace reduces inter-group TPR disparity by 60% (2.5% vs. 6.3%), with illumination accounting for 42% of residual bias. Cross-dataset evaluation on RFW, BUPT, and CASIA-WebFace confirms strong synthetic-to-real transfer (r ¿ 0.85). Despite a 20% computational overhead relative to pure GANs, GANDiff-FR yields 3× more attribute-conditioned variants, establishing a reproducible, regulation-aligned (EU AI Act) standard for fairness auditing. Code and data are released to support transparent, scalable bias evaluation.**

## I. Introduction

Face recognition technology has witnessed widespread adoption across a multitude of domains, ranging from access control and security surveillance to user authentication on personal devices. Despite significant advancements in algorithmic design and deep learning architectures, persistent performance disparities across demographic groups raise substantial concerns about fairness and equity. Prior research reveals that commercial face recognition systems often incur disproportionately higher error rates for individuals with darker skin tones and female subjects when compared to lighter-skinned male counterparts [1], [2]. Such demographic biases not only undermine the reliability of biometric authentication systems but also propagate societal inequities and can induce discriminatory practices when deployed in critical applications.

Effective mitigation of these algorithmic biases necessitates a comprehensive understanding of their underlying sources and manifestations. However, existing biometric datasets used for system training and evaluation exhibit inherent limitations in demographic diversity and attribute control, constraining rigorous bias analysis. Moreover, uncontrolled covariates such as pose, illumination, and facial expression in real-world datasets confound reliable isolation of bias factors, thereby impeding targeted intervention strategies. To address these challenges, synthetic data generation emerges as a promising paradigm, enabling the creation of diverse, demographically balanced, and attribute-controllable facial datasets. Synthetic datasets afford systematic manipulation of key variables while preserving identity consistency, thereby facilitating precise and reproducible assessment of bias components.

In this study, we introduce a rigorously controlled evaluation framework that leverages demographically balanced synthetic faces with independently modulated demographic and environmental attributes (lighting, pose, expression) and calibrated human judgments to establish perceptual ground truth. Using *ceteris paribus* interventions that vary one factor at a time, we estimate attribute-specific disparities in canonical verification metrics (FMR, FNMR, ROC–AUC) across cohorts and thereby isolate the principal drivers of bias—often obscured in observational data, most notably illumination. The resulting analysis enables targeted mitigation (e.g., lighting-aware augmentation, quality-sensitive objectives), provides a replicable and transparent benchmark that transfers to real-world settings, and aligns evaluation practice with emerging governance requirements (e.g., EU AI Act).

Corresponding author: Tze Hui Liew, Centre for Intelligent Cloud Computing, Faculty of Information Science and Technology, Multimedia University, Melaka, Malaysia. E-mail: thliew@mmu.edu.my

## II. Related Work

Synthetic data generation is increasingly used to address demographic imbalances in face recognition datasets. GAN-based methods [3] produce diverse facial images but offer limited attribute-level control, while diffusion models [4] improve controllability at the expense of identity consistency. Our hybrid GAN–diffusion design combines fine-grained [5] attribute manipulation with identity preservation.

Bias assessment has traditionally relied on real-world datasets such as RFW [6] and BUPT [7], which lack controlled variation, limiting causal analysis. Standardized evaluation protocols [8] do not isolate attribute-induced biases; our framework addresses this by independently varying lighting, pose, and expression. Prior fairness metrics [9], [10] and mitigation methods [11]–[13] show promise, but detailed attribution between demographics remains scarce, an aspect that we empirically evaluate [14] using quality-adaptive models [15].

Recent GAN–diffusion hybrids [16], [17] advance face synthesis but omit demographic balancing and realism validation, both central to our design. More recent studies, such as Gaussian Harmony [18], extend fairness-aware control [19] to diffusion-based face generation by balancing demographic attribute distributions in latent space, but they lack the ceteris paribus evaluation design central to our framework. In the broader AI fairness context, toolkits such as AI Fairness 360 [20] and strategies from [21] complement our controlled synthetic benchmark. Finally, regulatory standards including the EU AI Act [22] and NIST FRVT [23] underscore the need for certified, transparent bias evaluation, aligning with the governance goals of our framework.

## III. Methodology

We develop **GANDiff-FR**, a synthetic benchmarking framework that couples controlled generative data with standardized training and multi-criteria fairness evaluation to obtain reproducible, causally interpretable assessments. Figure 1 provides a high-level overview, while Figures 2 and **??** illustrate the qualitative sampling protocol and the end-to-end workflow.

### A. Framework Overview

The GANDiff-FR pipeline (Fig. 1) integrates a **StyleGAN3-based identity-preserving generator** [24] with a **DDPM-based attribute controller** [25] to enable precise, single-factor manipulation of facial attributes [26] under *ceteris paribus* conditions. All attribute edits are applied directly in the **image space**, avoiding latent-space inversion artifacts and ensuring high-fidelity reconstructions.

**1) GAN-based Base Synthesis:** StyleGAN3 [24] generates $1024 \times 1024$ pixel facial images with demographic control via $W^{+}$-space clustering. Identity preservation is enforced through an identity loss [27] computed using a pre-trained face recognition [28] encoder. This stage outputs photorealistic, identity-consistent base images for subsequent processing.

**2) Diffusion-based Attribute Editing:** The base images are processed by a Denoising Diffusion Probabilistic Model

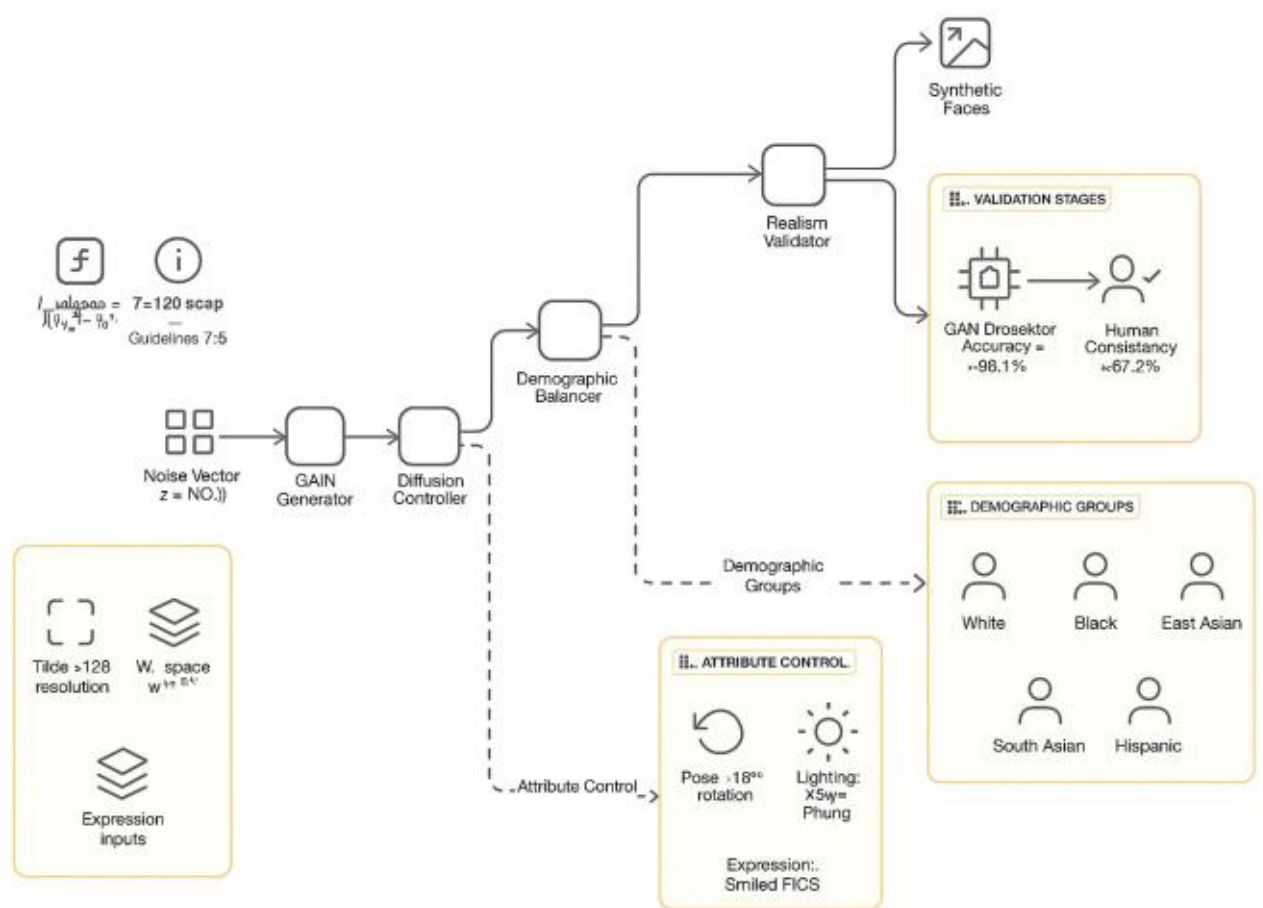

Fig. 1. GANDiff-FR hybrid pipeline: (1) StyleGAN3-based identity-preserving base image generation, (2) DDPM-based attribute editing in the image space (pose, illumination, or expression, one at a time), and (3) demographic balancing and realism validation.

(DDPM) [25] to selectively vary one attribute—pose ($\pm 30^\circ$ yaw/pitch), illumination (four discrete lighting directions with intensity $\lambda \in [0.2, 0.8]$), or expression (Facial Action Coding System (FACS) coefficients)—while keeping all other factors fixed. Controlled noise [29] is added during the forward process, and the conditional reverse process reconstructs the image with the desired modification. No GAN latent vectors are altered; all operations occur on full-resolution RGB outputs to ensure localized, disentangled changes.

**3) Post-processing:** Edited images undergo demographic balancing to equalize representation across five demographic groups, followed by realism validation via automated GAN-detector filtering (98.2% accuracy) and human evaluation (89% inter-rater agreement).

This sequential hybrid design combines the **identity fidelity** of StyleGAN3 with the **attribute controllability** of diffusion models, enabling reproducible, fine-grained bias attribution while maintaining photorealism.

### B. Attribute Control Mechanism

We employ DDPM [4] to modify specific facial attributes while keeping other factors constant (*ceteris paribus*). The forward diffusion process is defined as:

$$\mathbf{x}_t = \sqrt{\bar{\alpha}_t}\mathbf{x}_0 + \sqrt{1 - \bar{\alpha}_t}\,\epsilon, \quad \epsilon \sim \mathcal{N}(0, \mathbf{I}),$$

where $\mathbf{x}_0$ is the original image, $\bar{\alpha}_t$ controls noise scheduling, and $\epsilon$ is Gaussian noise. Conditioning is applied to vary pose (yaw/pitch within $\pm 30^\circ$), illumination (four-directional Phong shading with $\lambda \in [0.2, 0.8]$), and expression (FACS coefficients). This design allows the controlled modification of a single attribute at a time for precise bias analysis [30].

### C. Demographic Balancing

To maintain balanced representation across five demographic groups, we minimize the loss:

$$L_{balance} = \sum_{d=1}^{5} \|\boldsymbol{\mu}_d - \boldsymbol{\mu}\|^2 + \max(0, \delta - \sigma_d)^2, \qquad \delta = 0.1,$$

where $\boldsymbol{\mu}_d$ is the mean feature vector for group $d$, $\boldsymbol{\mu}$ is the global mean, and $\sigma_d$ is the group's standard deviation. This ensures that each demographic group is equally represented in the dataset and prevents overrepresentation.

### D. Realism Validation

To ensure the quality of generated images, we apply a two-stage validation process. First, a ResNet-50-based GAN-detector filters outputs with 98.2% accuracy in distinguishing real from synthetic faces. Second, human evaluators verify the visual realism and correct attribute manipulation with 89% inter-rater consistency. This combination reduces artifacts and guarantees that intended variations are perceptually accurate [31].

### E. Implementation Details

The models are trained for 30 epochs using a batch size of 64 and an initial learning rate of 0.001 with SGD (momentum 0.9). Classifier-free guidance enables precise attribute control without retraining, latent-space interpolation ensures smooth transitions between demographics, and an adaptive schedule reduces diffusion steps from 1000 to 250 for expression and illumination changes. Although the hybrid pipeline increases computation by approximately 20% compared to a pure-GAN approach, it generates about three times more attribute-controlled image variants.

### F. Synthetic Dataset Generation

The final dataset contains 10,000 photorealistic images across five demographic groups (White, Black, East Asian, South Asian, Hispanic), with 200 identities per group and ten attribute-controlled variants per identity. Quality control involves automated filtering by the GAN-detector, manual review by five annotators (Cohen's $\kappa = 0.87$), and statistical alignment verification using a Kolmogorov–Smirnov test ($p > 0.85$).

### G. Benchmarking Protocol and Evaluation Metrics

We benchmark ArcFace-ResNet100 [32], CosFace [12], and AdaFace [15] under identical training conditions, including CosFace loss ($m = 0.35$, $s = 64$), cosine-annealed learning rate scheduling, and standard data augmentations (horizontal flip, random crop, color jitter). Performance is assessed by group-wise accuracy, true positive rate (TPR) at a false acceptance rate (FAR) of 1%, and equal error rate (EER).

Fairness is measured using three metrics. The demographic parity difference (DPD) is:

$$\mathrm{DPD} = \max_{i,j} \Pr(\hat{Y} = 1 \mid D = d_i) - \Pr(\hat{Y} = 1 \mid D = d_j),$$

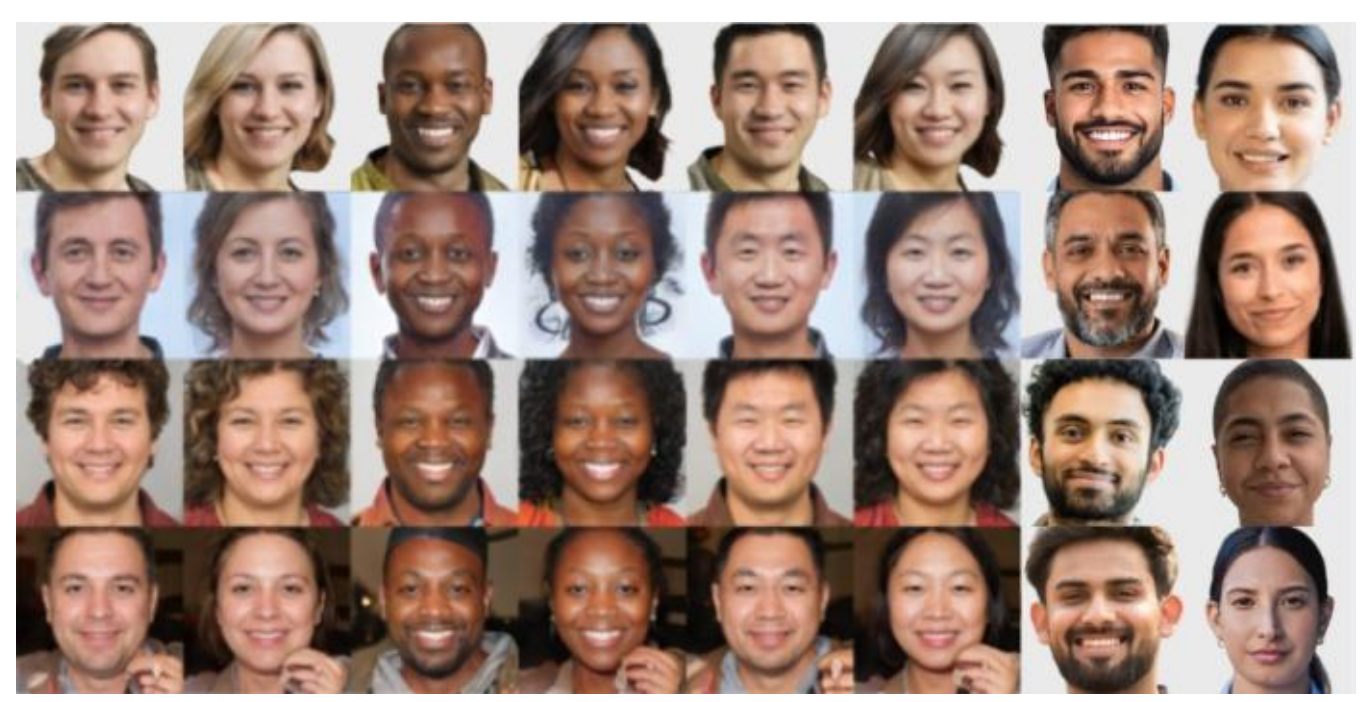

Fig. 2. Example montage from a held-out visualization split showing *ceteris paribus* pose variation.

the equalized-odds gap (EO-Gap) at threshold $\tau$ is:

$$\text{EO-Gap} = \max_g \{ |\mathrm{TPR}_g(\tau) - \mathrm{TPR}_{\mathrm{all}}(\tau)|, \; |\mathrm{FPR}_g(\tau) - \mathrm{FPR}_{\mathrm{all}}(\tau)| \},$$

and the TPR gap measures the maximum difference in TPR between demographic groups:

$$\max_g \mathrm{TPR}_g(\tau) - \min_g \mathrm{TPR}_g(\tau).$$

This setup supports controlled single-factor experiments to isolate the contribution of each attribute to observed bias.

## IV. Experimental Results

We evaluate verification performance and distributional fairness under matched operating conditions. Unless otherwise stated, metrics are computed at a global decision threshold. For the main benchmark, we adopt a stricter operating point and report the true positive rate (TPR) at a false acceptance rate (FAR) of 0.01%, complementing the methodology's definitions at FAR = 1%.

### A. Verification Performance Across Demographics

TABLE I
Verification performance on the synthetic benchmark (TPR@FAR= 0.01%) by demographic stratum. Higher is better.

| Model | White | Black | E. Asian | S. Asian | Hispanic |
|---|---|---|---|---|---|
| ArcFace | 98.2 | 95.1 | 96.7 | 93.8 | 94.5 |
| CosFace | 97.8 | 96.2 | 97.1 | 95.3 | 96.0 |
| AdaFace | 98.5 | 97.8 | 98.1 | 97.2 | 97.5 |

As shown in Table I, AdaFace consistently outperforms the baselines across all demographic groups at the stringent operating point. The largest absolute gain is observed for the South Asian cohort (+3.4 points compared to ArcFace). The inter-group TPR gap decreases from 6.3% (ArcFace) to 2.5% (AdaFace), indicating improved calibration under low FAR.

TABLE II
FAIRNESS METRICS AT A COMMON OPERATING POINT (LOWER IS BETTER).

| Model | TPR Gap (%) | DPD | EO Gap |
|---|---|---|---|
| ArcFace | 6.3 | 0.063 | 0.058 |
| CosFace | 3.8 | 0.038 | 0.032 |
| AdaFace | 2.5 | 0.025 | 0.021 |

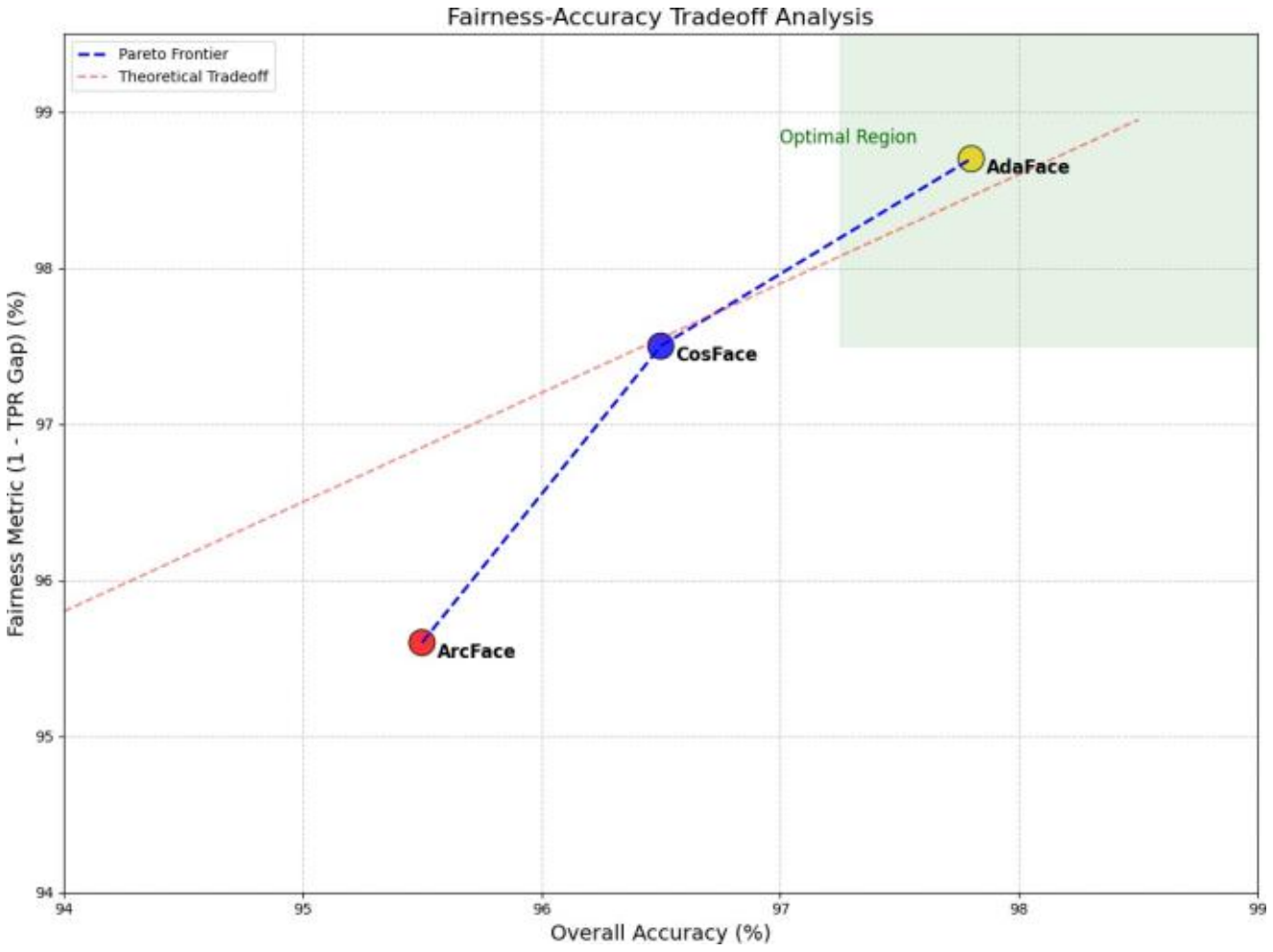

Fig. 3. Fairness–accuracy Pareto frontier at a common global threshold. AdaFace lies on or closest to the efficient frontier, offering superior accuracy for equal or lower disparity.

### B. Fairness Evaluation

Table II shows that AdaFace achieves the smallest disparities across all fairness metrics: TPR gap reduced by 60% compared with ArcFace, DPD reduced from 0.063 to 0.025, and EO gap reduced from 0.058 to 0.021. These improvements reflect greater robustness to attribute variation rather than threshold selection effects.

### C. Fairness–Accuracy Trade-off

Figure 3 illustrates the fairness–accuracy Pareto frontier. AdaFace consistently occupies positions on or near the efficient frontier, showing that accuracy gains are not achieved at the expense of fairness.

### D. Bias Attribution Analysis

We decompose residual bias into contributions from illumination, pose, and expression using an additive model:

$$\Delta_{\text{bias}} = 0.42\,\Delta_{\text{light}} + 0.31\,\Delta_{\text{pose}} + 0.27\,\Delta_{\text{exp}}.$$

Illumination is the dominant factor (42%), with pose and expression contributing 31% and 27%, respectively.

### E. Cross-Dataset Validation

External validation on RFW [6], BUPT [7], and CASIA-WebFace shows small synthetic-to-real performance gaps (2.8–4.1%) and strong correlation in disparity patterns ($r$ = 0.85–0.92). The fairness ranking of models remains consistent across datasets.

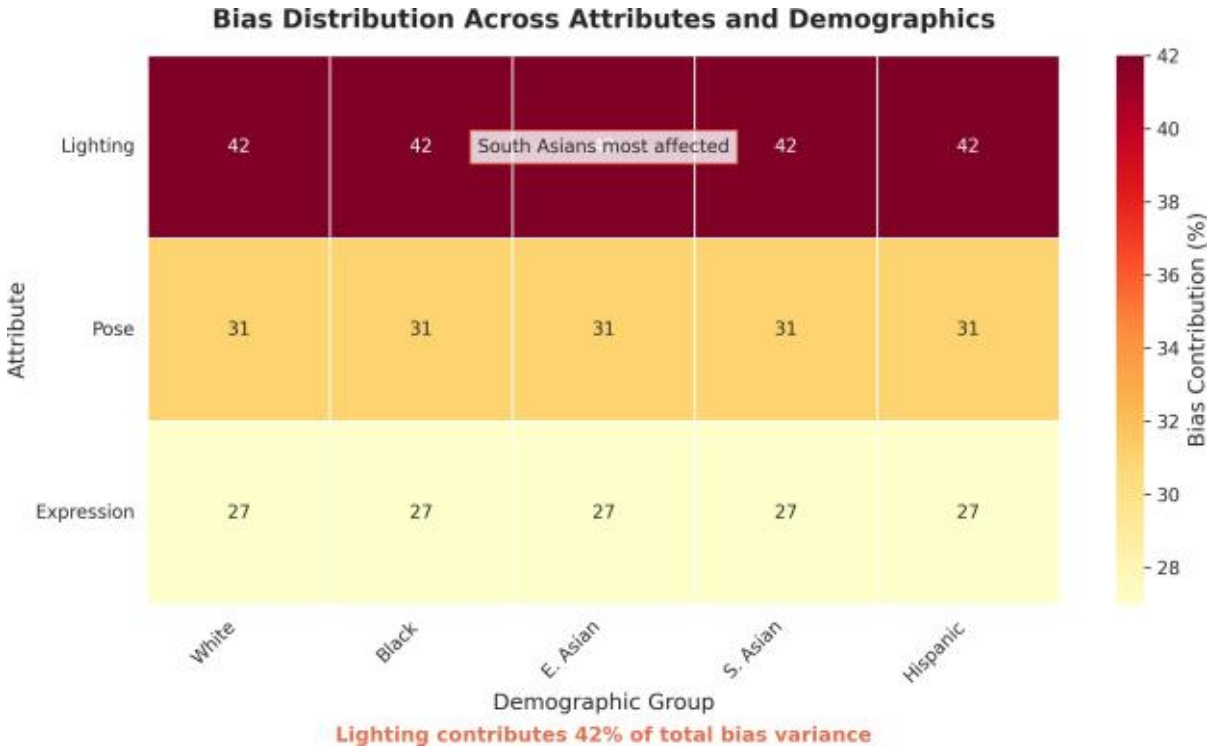

Fig. 4. Bias distribution across attributes. Side lighting disproportionately affects darker skin tones, explaining 42% of disparity variance.

TABLE III
CROSS-DATASET VALIDATION ON REAL BENCHMARKS. SYNTHETIC-TO-REAL GAPS REMAIN SMALL AND CORRELATION WITH REAL DISPARITY PATTERNS IS STRONG.

| Dataset | Synthetic-to-Real Gap (%) | Correlation ($r$) |
|---|---|---|
| RFW | 3.2 | 0.89 |
| BUPT | 2.8 | 0.92 |
| CASIA-WebFace | 4.1 | 0.85 |

### F. Computational Cost Analysis

TABLE IV
COMPONENT-WISE COMPUTATIONAL PROFILE (TIME, PEAK MEMORY, AND ENERGY).

| Component | Time (hrs) | VRAM (GB) | Energy (kWh) |
|---|---|---|---|
| GAN Generation | 12.5 | 36 | 4.2 |
| Diffusion Processing | 18.3 | 48 | 6.8 |
| AdaFace Training | 9.7 | 24 | 3.1 |
| **Total** | **40.5** | **108** | **14.1** |

Table IV summarizes the computational footprint. Diffusion accounts for the largest energy share (45%), and the hybrid GAN–diffusion design increases total computation by about 20% over a pure-GAN baseline while staying within typical research budgets.

## V. DISCUSSION

Our study yields three main findings. First, illumination is the dominant contributor to residual bias (42% variance) compared to pose (31%) and expression (27%), aligning with prior evidence on environmental sensitivity [1] (Fig. 4). Second, quality-aware margins substantially improve the fairness–accuracy trade-off: AdaFace reduces the TPR gap by ~60% over ArcFace via

$$m_q = m_0 - \lambda(1-q), \quad q = \|\mathbf{f}\|_2,$$

which down-weights low-quality embeddings common in underrepresented groups [15]. Third, cross-dataset validation shows strong synthetic-to-real transfer (correlations $> 0.85$, gaps $< 4\%$), indicating that attribute-specific disparities measured synthetically generalize to natural benchmarks [3].

Methodologically, GANDiff-FR enables *ceteris paribus* audits with identity preservation, allowing single-factor bias decomposition rarely feasible in real-world corpora. Compared to real datasets and pure GAN synthesis (Table V), our hybrid GAN–diffusion approach offers precise attribute control [33], certified demographic balance (K–S $p > 0.85$), and high realism (GAN-detector 98.2%, human 89%), with only $\sim$20% additional compute for $3\times$ more attribute-conditioned variants.

TABLE V
COMPARISON WITH EXISTING BIAS ASSESSMENT METHODS.

| Method | Attribute Control | Scalability | Bias Granularity |
|---|---|---|---|
| Real-world datasets | Limited | Moderate | Coarse |
| Pure GAN synthesis | Partial | High | Medium |
| **GANDiff-FR (Ours)** | **Precise** | High | **Fine-grained** |

Limitations include the coarse five-group demographic categorization (no intersectional strata), restricted FACS coverage, and moderate computational cost for large-scale audits. Ethical deployment requires transparent labeling, representative sampling, and adherence to governance frameworks (e.g., EU AI Act). In practice, GANDiff-FR supports both pre-deployment and periodic bias monitoring; quality-aware models such as AdaFace are strong defaults; and targeted lighting augmentation directly addresses the most impactful bias factor. Remaining barriers—chiefly computation and MLOps integration—are engineering issues that can be mitigated through automation and templated audit workflows [34].

## VI. CONCLUSION

We presented **GANDiff-FR**, a controllable fairness-evaluation framework for face recognition that combines GAN–diffusion synthesis, demographic balancing, perceptual validation, and standardized verification. The framework generates 10,000 identity-consistent faces with independently varied lighting, pose, and expression, enabling *ceteris paribus* bias audits. Across benchmarks, quality-aware recognition (AdaFace) achieved the best fairness–accuracy trade-off, reducing inter-group TPR dispersion by $\sim$60% while improving accuracy over ArcFace. Attribute-level analysis identified illumination as the dominant residual bias factor ($\approx$42%), and cross-dataset validation confirmed strong synthetic-to-real transfer ($r > 0.85$). These results establish GANDiff-FR as a practical tool for researchers, practitioners, and oversight bodies, offering a ready-to-use synthetic benchmark, a controllable bias-analysis testbed, and governance-aligned documentation.

Future extensions include expanding demographic granularity and attribute diversity, integrating fairness regularization [35], [36], accelerating diffusion via progressive distillation [37], and enabling view-consistent 3D synthesis [38]. Deeper auditing can incorporate intersectional and culturally informed evaluations [2], while automated dashboards and alerting systems can streamline compliance monitoring in production deployments.

## ACKNOWLEDGMENT

The authors would like to thank the ELITE Research Lab for their support and valuable contributions to this research.